\pdfoutput=1
\documentclass[11pt]{article}

\usepackage[]{acl}

\usepackage{times}
\usepackage{latexsym}

\usepackage[T1]{fontenc}
\usepackage[utf8]{inputenc}

\usepackage{microtype}

\usepackage{amsmath, adjustbox}
\usepackage{graphicx}
\usepackage[normalem]{ulem}
\useunder{\uline}{\ul}{}

\title{Using contradictions improves question answering systems}

\author{
Étienne Fortier-Dubois
\And 
Domenic Rosati \\ scite.ai \\ Dalhousie University
}

\begin{document}
\maketitle
\begin{abstract}
This work examines the use of \emph{contradiction} in natural language inference (NLI) for question answering (QA). Typically, NLI systems help answer questions by determining if a potential answer is \emph{entailed} (supported) by some background context. But is it useful to also determine if an answer contradicts the context? We test this in two settings, multiple choice and extractive QA, and find that systems that incorporate contradiction can do slightly better than entailment-only systems on certain datasets. However, the best performances come from using contradiction, entailment, and QA model confidence scores together. This has implications for the deployment of QA systems in domains such as medicine and science where safety is an issue.

% Ensuring the safety of question answering (QA) systems is critical to their deployment in medicine and science. One approach to improving these systems uses natural language inference (NLI) to determine whether answers are supported, or entailed, by some background context. 
%However, these systems are vulnerable to supporting an answer with a source that is wrong or misleading. 
% Our work adds contradiction to entailment: answers are ranked better when they are not contradicted by the context. We evaluate this system on multiple choice and extractive QA and find that while the contradiction-based systems are competitive with and sometimes better than entailment-only systems, models that incorporate contradiction, entailment, and QA model confidence scores together are the best. 
%Based on this result, we explore unique opportunities for leveraging contradiction-based approaches such for improving interpretability and selecting better answers.

\end{abstract}

\section{Introduction}

Safety in NLP systems is unresolved, particularly in biomedical and scientific contexts where hallucination, overconfidence, and other problems are major obstacles to deployment \citep{ji_survey_2022,kell_what_2021}. One active area of research to solve these issues is natural language inference (NLI) \citep{li_faithfulness_2022}. NLI is the task of determining whether a hypothesis is true (\emph{entailed}), false (\emph{contradicted}), or undetermined (\emph{neutral}) given some premise. 

\begin{figure}
    \centering
  \includegraphics[width=1.0\linewidth]{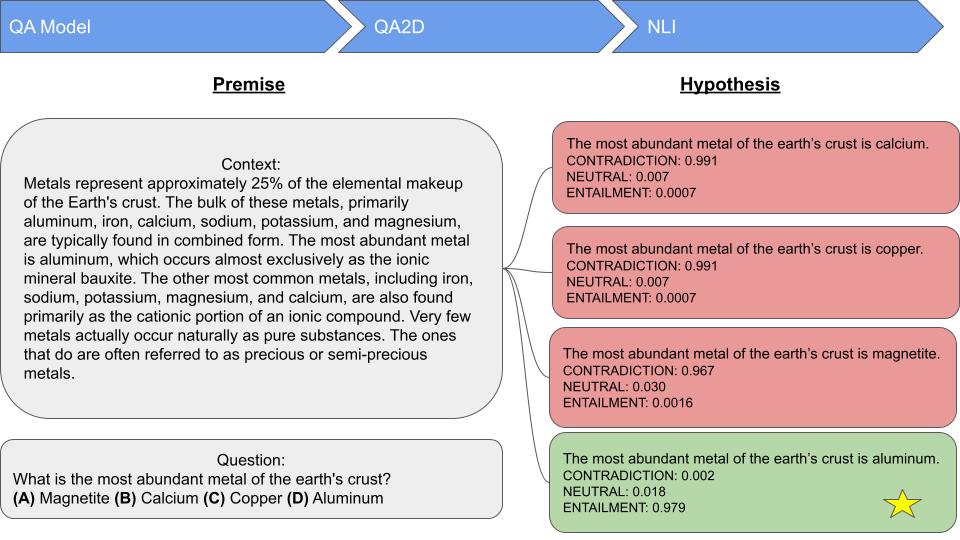}
  \caption{A QA model is used to produce answers which are reformulated as hypotheses to determine if they are entailed or contradicted by a premise. The answers are ranked by NLI class scores to select the best answer.}
  \label{fig:architecture}
\end{figure}

Current NLI systems typically focus only on entailment to verify hypotheses---they calculate the degree to which a hypothesis is supported by the premise. But the premise can provide another signal: contradiction. Regardless of how well a hypothesis is entailed by the context, it can also be more or less contradicted, which could affect whether it is accepted or rejected. Contradictions are an important signal indicating whether some statement might be unacceptable given a premise. In some cases where we might not know if a statement is supported, we should still ensure we are rejecting statements that are outright contradicted.

We wondered if adding this signal to a question answering (QA) system might improve performance and safety. To this end, we propose a method that reformulates answers from the QA system as hypotheses for NLI, calculates the entailment, contradiction, and neutrality of each hypothesis, and then selects the best one based on a combination of these results
(Figure \ref{fig:architecture}). We show that across 16 QA datasets (9 multiple choice and 7 extractive), the best approach is to use entailment, contradiction, and confidence scores together. Using only contradiction is roughly on par with, and sometimes better than, using only entailment.

\subsection{Related work}

\textbf{NLI for question answering} has been explored by several authors in various settings; see \citet{paramasivam_survey_2021} for an overview. 

One of these settings is \textbf{selective question answering for extractive QA}, where \emph{selective} refers to abstention when the system is not confident enough in its answer \citep{kamath_selective_2020}. \citet{chen_can_2021} have found that NLI systems are able to verify the predictions made by a QA system in this setting, but their result is limited to only selecting a top $k\%$ of answers. Moreover, they do not provide an approach for improving overall performance, nor do they show the effect of incorporating contradiction directly (but do so indirectly by analyzing non-entailed passages).

In the related setting of \textbf{multiple choice QA and fact checking}, \citet{mishra_looking_2021} have explored the use of entailment, finding that NLI models do well at these tasks by themselves, but can perform even better when they are adapted to in-domain data and longer premises. Yet their method uses only a two-class NLI set up (entailed or not entailed), which doesn't tell us much about directly using the contradiction signal. \citet{pujari-goldwasser-2019-using} does incorporate the contradiction signal showing the power of contradiction to improve machine comprehension but does not analyze its effects separately from entailment.

Other QA settings in which NLI has been used include open domain \citep{harabagiu_methods_2006} and multi-hop \citep{trivedi_repurposing_2019}. Thus far, approaches tend to focus on entailment. To our knowledge, our work is the first to directly assess using contradictions for QA isolated from entailment.

Outside of question answering, a domain that uses contradictions is \textbf{factual consistency}---the task of ensuring that a collection of utterances is faithful to a source document. \citet{li_faithfulness_2022} provide an overview. Typically, entailment is still the main focus, but \citet{laban_summac_2022} propose an NLI-based method to ensure the consistency of a summary with a source document using contradiction and neutral scores in addition to entailment, beating out previous systems.

Other researchers have used contradictions to identify consistency errors across Wikipedia \citep{schuster_stretching_2022, hsu_wikicontradiction_2021} or generate credible character dialogue \citep{nie_i_2021, song_generating_2020}. 

\section{Methods}

We tested the effect of contradictions in two QA settings and a total of sixteen question-answer datasets. Our approach is broadly similar to both \citet{chen_can_2021} and \citet{mishra_looking_2021} in that we use most of the same datasets for evaluating NLI reranking for multiple choice QA and extractive QA. Unlike both, we incorporate contradiction directly as a signal for reranking answers.

Briefly, for each dataset, we used pretrained QA models to produce answers and confidence scores for the dataset's questions. We refer to the confidence scores below as \textbf{QA}. We then trained QA2D models (where QA2D stands for "question-answer to declarative") to turn the answers into the declarative hypothesis format required for NLI. For example, the question-answer pair "What is the most abundant metal in the Earth crust? Copper." might be rephrased as "The most abundant metal in the Earth crust is copper" (see Appendix \ref{sec:appendix_qa2d_models} for more details).

With the question contexts as premises, we then used NLI models to classify every premise-hypothesis pair into three classes, each with an associated score: entailed (\textbf{E}), contradicted (\textbf{C}), and neutral (\textbf{N}). After that, we trained logistic regression calibration models to find which linear combination of the four scores---\textbf{QA}, \textbf{E}, \textbf{C}, and \textbf{N}---was best able to pick the answers accurately.

When evaluating performance, we applied the selective QA approach from \citet{kamath_selective_2020} to rank answers using combinations of the four scores, and then consider only those that the model was most confident in answering. We compared selecting the top 20\% and 50\%. In the multiple choice setting, it was also possible to rank all potential answers according to the four scores, unlike in the extractive QA setting where the QA model produced only one answer per question, so we evaluated performance with that approach as well (see appendix \ref{sec:appendix_answer_ranking} for details).

\section{Experimental setting}

In the multiple choice setting, we tested 9 datasets. Two of them are in-domain, since the pretrained QA models we used were finetuned on them. Specifically, we used a RoBERTa large model \citep{liu_roberta_2019} finetuned on the RACE dataset \citep{lai_race_2017}, as well as two DeBERTa v3 variants, base and xsmall \citep{he_debertav3_2021}, finetuned on the SciQ dataset \citep{welbl_crowdsourcing_2017}.

In the extractive QA setting, we used 7 datasets: five from the MRQA 2019 task \citep{fisch_mrqa_2019}, as well as SQuAD 2.0 \citep{rajpurkar_know_2018} and SQuAD adversarial \citep{jia_adversarial_2017}. The SQuAD model is the in-domain dataset: it was used to pretrain \citep{rajpurkar_squad_2016} the two QA models we used, DistillBERT \citep{sanh_distilbert_2020} and BERT-Large \citep{devlin_bert_2019}. Like \citet{chen_can_2021}, we used the Natural Questions dataset for calibration.

In both settings, all datasets contain the relevant context that can be used by the QA models to select answers. More detail on the datasets and QA models is available in appendices \ref{sec:appendix_datasets} and \ref{sec:appendix_qa_models} respectively. 

See appendices \ref{sec:appendix_qa2d_models}, \ref{sec:appendix_nli_models}, and \ref{sec:appendix_calibration_models} for details on the QA2D, NLI, and calibration models. Our models follow the setups described in \citet{kamath_selective_2020}, \citet{chen_can_2021}, and \citet{mishra_looking_2021}. The main interesting detail is that the calibration models were trained on a holdout set of 100 samples from a single domain, using logistic regression, as in \citet{chen_can_2021}.

\section{Results}

\subsection{Multiple choice setting}

\begin{table*}[]
\centering
\begin{tabular}{lllllllllll}
\hline
{\small QA Model} & {\small Cosmos} &  {\small DREAM} & {\small MCS} & {\small MCS2} & {\small MCT} & {\small QASC} & {\small RACE} & {\small RACE-C} & {\small SciQ} & {\small \emph{Average}} \\ \hline
SciQ-base & 18.46 & 43.80 & 61.99 & 63.71 & 44.76 & 93.41 & 30.97 & 27.39 & 95.28 & 53.30 \\
SciQ-small & 25.46 & 48.26 & 60.28 & 66.04 & 59.76 & 90.60 & 35.56 & 30.62 & 98.09 & 57.18 \\
QA & 64.22 & 82.56 & 89.70 & 86.98 & 90.48 & 98.16 & 76.93 & \textbf{69.80} & 97.96 & 84.08 \\ \hline
QA+E+C & {\ul 64.72}* & \textbf{83.19}* & \textbf{90.06}* & \textbf{87.59}* & \textbf{91.43}* & \textbf{98.60} & \textbf{77.53}* & \textbf{69.80}* & \textbf{98.21} & \textbf{84.57} \\
QA+E & 64.32 & {\ul 82.85}* & {\ul 89.92}* & {\ul 87.29}* & {\ul 91.07} & {\ul 98.49*} & {\ul 77.18} & 69.66 & {\ul 98.09} & 84.31 \\
QA+C & \textbf{64.82} & 82.75* & 89.88* & {\ul 87.29}* & 90.83 & 98.38 & 77.16 & \textbf{69.80} & {\ul 98.09} & {\ul 84.33} \\ \hline
\end{tabular}
\caption{\emph{Multiple choice setting}. Accuracy scores (best per column in \textbf{bold}, second best {\ul underlined}, statistical significance (pairwise students t-test) is indicated by asterix) after answer ranking with the mnli-large NLI model. The top three rows show the accuracy of using only the QA models' confidence score; "QA" refers to the scores of the RoBERTa-RACE model, which was used for calibration. The bottom rows add the entailment and/or contradiction scores to the RoBERTa-RACE score. For other NLI models, and for just E, C, and E+C without calibration with RoBERTa-RACE, see Table \ref{tab:cross_mc_performance} in the appendix.}
\label{tab:calibrated_performance}
\end{table*}

For most multiple choice datasets, the best accuracy---when ranking all potential answers---is attained when using a calibrated model combining QA confidence, entailment, and contradiction (\textbf{QA+E+C} in Table \ref{tab:calibrated_performance}). Only for the in-domain case (RACE-C) does the uncalibrated RoBERTa-RACE model perform on par with that. Using QA scores combined with either entailment (\textbf{QA+E}) or contradiction (\textbf{QA+C}) achieves similar performance, with contradiction winning by a small margin: 84.33\% average accuracy compared to 84.31\%. 

To inspect these trends further, we performed a correlation analysis of the NLI classes and QA confidence scores with the correct answer (appendix \ref{sec:appendix_correlation_analysis}). We found that besides QA confidence, it is the contradiction score that has the strongest correlation with the correct answer. The analysis also showed that the neutral class score (\textbf{N}) had almost no effect, which is why it is omitted in all results.

\begin{table*}[]
\centering
\begin{tabular}{lllllllll}
\hline
 & Dataset & QA +E+C & QA+C & QA+E & E+C & E & C & QA \\ \hline
20\% & CosmosQA & 77.55 & \textbf{91.12} & 76.88 & 69.18 & 68.34 & 83.25 & {\ul 88.61} \\
 & DREAM & {\ul 98.28} & \textbf{98.77} & {\ul 98.28} & 96.32 & 96.32 & 96.81 & {\ul 98.28} \\
 & MCScript & \textbf{99.82} & 99.46 & \textbf{99.82} & {\ul 99.64} & {\ul 99.64} & 99.46 & \textbf{99.82} \\
 & MCScript-2.0 & {\ul 99.58} & \textbf{99.72} & 99.45 & 99.17 & 99.03 & 97.37 & {\ul 99.58} \\
 & MCTest & \textbf{100} & {\ul 99.40} & \textbf{100} & \textbf{100} & \textbf{100} & {\ul 99.40} & 98.81 \\
 & QASC & \textbf{100} & \textbf{100} & \textbf{100} & \textbf{100} & \textbf{100} & \textbf{100} & \textbf{100} \\
 & RACE & 94.93 & {\ul 96.69} & 94.72 & 92.44 & 92.24 & 90.17 & \textbf{98.24} \\
 & RACE-C & 88.73 & {\ul 92.96} & 89.44 & 85.21 & 85.92 & 86.62 & \textbf{93.66} \\
 & SciQ & \textbf{100} & \textbf{100} & \textbf{100} & \textbf{100} & \textbf{100} & \textbf{100} & \textbf{100} \\
 & \emph{Average} & 95.43 & \textbf{97.57} & 95.40 & 93.55 & 93.50 & 94.79 & {\ul 97.45} \\ \hline
50\% & CosmosQA & {\ul 80.29} & \textbf{81.70} & 76.94 & 75.80 & 70.64 & 80.63 & 76.47 \\
 & DREAM & 95.10 & \textbf{96.86} & 94.90 & 93.63 & 93.63 & 93.63 & {\ul 96.67} \\
 & MCScript & 98.57 & {\ul 98.64} & 98.28 & 98.00 & 97.93 & 97.14 & \textbf{98.78} \\
 & MCScript-2.0 & 96.40 & \textbf{98.23} & 95.84 & 94.68 & 94.40 & 96.01 & {\ul 98.01} \\
 & MCTest & 99.52 & \textbf{99.76} & 99.52 & 99.05 & 99.05 & {\ul 99.76} & 99.52 \\
 & QASC & \textbf{100} & \textbf{100} & \textbf{100} & {\ul 99.78} & {\ul 99.78} & {\ul 99.78} & \textbf{100} \\
 & RACE & 90.11 & {\ul 92.68} & 89.99 & 87.71 & 87.38 & 85.23 & \textbf{93.88} \\
 & RACE-C & 85.11 & 84.83 & {\ul 85.39} & 78.37 & 78.37 & 77.25 & \textbf{87.36} \\
 & SciQ & \textbf{100} & \textbf{100} & \textbf{100} & \textbf{100} & \textbf{100} & 99.74 & \textbf{100} \\
 & \emph{Average} & 93.90 & \textbf{94.74} & 93.43 & 91.89 & 91.24 & 92.13 & {\ul 94.52} \\ \hline
\end{tabular}
\caption{\emph{Multiple choice setting}. Accuracy scores (best per row in \textbf{bold}, second best {\ul underlined}) for selective QA with 20\% and 50\% coverage of the dataset. Calibrations and QA confidence are all from RoBERTa-RACE, where RACE is the in-domain dataset.}
\label{tab:selective_mc_qa}
\end{table*}

When using the selective QA approach and evaluating only the 20\% of 50\% most confident answers, the best performance is attained with the \textbf{QA+C} combination (Table \ref{tab:selective_mc_qa}). This model is the only one that beats just using the QA confidence score on average. It is stronger than \textbf{QA+E+C} and \textbf{QA+E} for both coverage percentages.

Contradiction alone, without QA confidence scores (\textbf{C}), also beats both entailment alone (\textbf{E}) and entailment with contradiction (\textbf{E+C}) for both coverages. These results match our intuition that the less contradicted an answer, the more likely it is correct, even in cases where there is uncertainty about its entailment.

\subsection{Extractive QA setting}

\begin{table*}[]
\centering
\begin{tabular}{lllllllll}
\hline
& Dataset & QA+E+C & QA+C & QA+E & E+C & E & C & QA \\ \hline
20\% & BioASQ & {\ul 85.04} & 83.10 & \textbf{85.06} & 74.22 & 74.22 & 75.47 & 82.99 \\
 & HotpotQA & {\ul 86.62}  & 85.89 & \textbf{86.69} & 80.60 & 80.60 & 79.82 & 85.33 \\
 & Natural Questions & {\ul 91.84}  & \textbf{92.18} & 91.68 & 79.89 & 79.87 & 82.09 & 90.98 \\
 & SQuAD & 98.26 & {\ul 98.76} & 92.37 & 98.17 & 92.48 & 90.88 & \textbf{99.04} \\
 & SQuAD-adv & \textbf{43.99}  & 43.57 & {\ul 43.98} & 43.74 & 43.60 & 42.81 & 39.83 \\
 & SQuAD2 & {\ul 37.64}  & 36.07 & 37.56 & 37.43 & 37.31 & \textbf{37.68} & 30.52 \\
 & TriviaQA & \textbf{81.33}  & 80.36 & 81.21 & 65.53 & 65.25 & 69.13 & 80.68 \\
 & \emph{Average} & {\ul 74.96}  & 74.19 & \textbf{74.99} & 67.68 & 67.62 & 68.27 & 72.77 \\ \hline
50\% & BioASQ & \textbf{76.13}  & 75.51 & {\ul 76.04} & 71.49 & 71.49 & 72.97 & 75.49 \\
 & HotpotQA & \textbf{79.37} & 78.95 & {\ul 79.30}  & 77.43 & 77.43 & 77.31 & 78.74 \\
 & Natural Questions & \textbf{84.53}  & 83.24 & {\ul 84.48} & 74.96 & 74.93 & 78.62 & 82.47 \\
 & SQuAD & {\ul 96.98}  & 97.01 & 96.97 & 91.58 & 91.52 & 91.19 & \textbf{97.00} \\
 & SQuAD-adv & 41.80  & 41.49 & 41.16 & {\ul 42.76} & \textbf{42.79} & 42.03 & 40.26 \\
 & SQuAD2 & 29.41 & 28.77 & 28.45 & \textbf{34.43} & 34.14 & {\ul 34.39} & 26.18 \\
 & TriviaQA & 74.30 & 74.23 & \textbf{74.37} & 65.05 & 64.93 & 68.08 & 74.21 \\
 & \emph{Average} & \textbf{68.93} & 68.46 & {\ul 68.68} & 65.39 & 65.32 & 66.37 & 67.76 \\ \hline
\end{tabular}
\caption{\emph{Extractive QA setting}. F1 scores (best per row in \textbf{bold}, second best {\ul underlined}) for selective QA with 20\% and 50\% coverage of the dataset. Calibrations and QA confidence are from the BERT-large model, where SQuAD is the in-domain dataset. For similar results on the smaller DistillBERT model, see Table \ref{tab:selective_distillbert} in the appendix.}
\label{tab:selective_extractive_qa}
\end{table*}

Similar results occur when evaluating the extractive QA datasets with 20\% and 50\% selective coverage (Table \ref{tab:selective_extractive_qa}). The \textbf{QA+C} model does better than \textbf{QA} alone, and \textbf{C} alone does better than \textbf{E+C} or \textbf{E} alone, indicating the importance of the contradiction signal here too. However, entailment seems to matter more for extractive QA, as the best F1 score overall was from \textbf{QA+E} in the 20\% coverage case, and \textbf{QA+E+C} in the 50\% case. 

\section{Discussion}

Contradiction with background context is a useful signal that NLP systems can use to infer answers to questions. This is not necessarily a superior strategy to using entailment, but our results show that combining these two signals can improve performance beyond what QA models can achieve on their own.  These results are interesting because using contradictions comes with potential benefits for the safety of NLP systems and, as a result, their deployment in domains such as medicine or science. Namely, that there are many potential cases where we are not sure if a statement is entailed directly by a background context but we may be sure that the statement is not refuted by a background context. In two-class NLI settings where we focus only on entailment, neutral and contradiction are collapsed together and we don't have this guarantee.

\section{Limitations}

Our work comes with some limitations. It is uncertain whether our results in two specific settings, multiple choice and extractive QA, would extend to more general settings for NLI, although the use of contradictions for factual consistency by \citet{laban_summac_2022} suggests that they could. Additionally, 3-class NLI is not sufficient to capture all the natural language relations that might be needed to verify an answer. As such more challenging datasets in other settings and more granular NLI settings should be attempted.

Another limitation involves answer ranking and the associated computational cost. The main reason we did not test answer ranking in extractive QA is that we did not generate diverse outputs, but another reason is that such a procedure grows prohibitively expensive as the domain becomes more open. In a fully open domain, ranking would require a quadratic evaluation for each context passage against each reformulated answer candidate \citep{schuster_stretching_2022}. Future work should look at comparison approaches that amortize this cost, such as NLI-based dense passage retrieval \citep{reimers_sentence-bert_2019}.

\bibliography{anthology,custom}
\bibliographystyle{acl_natbib}

\appendix

\section{Answer ranking procedure}
\label{sec:appendix_answer_ranking}

In the multiple choice setting, we performed an answer ranking procedure to pick the answer to a given question among a set of alternative answers $N$, using both NLI class scores and QA confidence scores. (This is distinct from the selection procedure for the top 20\% or 50\% of answers we used in both settings.) 

Similar to \citet{harabagiu_methods_2006}, answers are ranked based on the highest probability from the calibration model $\sigma$, given a linear combination of the QA or NLI scores given an answer $n \in N$ answer set. When a single feature is used, such as an NLI class or the QA score, no calibration is made and $\sigma$ is simply the identity of the confidence score. In the case of contradiction only, $\sigma$ is the inverse of the contradiction confidence score, indicating the least contradicted answer is being selected. Formally, our approach can be described as:
$$
\underset{N}{\operatorname{argmax}}\:\sigma(\text{QA}_n;\text{NLI}_n)
$$
where $\text{QA}_n$ is the QA model confidence score for answer $n$, and $\text{NLI}_n$ represents the various NLI class scores for $n$.

We did not use this approach in extractive QA, because we found that asking the model for the top $K=4$ answer produced almost the same four answer alternatives with slightly different spans each time. 

\section{Datasets}
\label{sec:appendix_datasets}

Tables \ref{tab:mc_qa_datasets} (multiple choice) and \ref{tab:extractive_qa_datasets} (extractive QA) outline the datasets we used. Additional details such as train size and preprocessing steps are available in the references provided. When space doesn't allow CosmosQA is aliased to Cosmos, MCScript to MCS, MCScript-2.0 to MCS2, and MCTest to MCT. The only preprocessing step we performed was to filter out questions where no context passage is provided. Validation splits (as opposed to test splits) are used in the CosmosQA and QASC cases, since context passages or gold standard answers are not available for these datasets.

\begin{table*}[t!]
\centering
\begin{tabular}{llll}
\hline
Dataset & Split & Size & Reference \\ \hline
CosmosQA & validation & 2985 & \citet{huang_cosmos_2019} \\
DREAM & test & 2041 & \citet{sun_dream_2019} \\
MCScript & test & 2797 & \citet{ostermann_mcscript_2018} \\
MCScript-2.0 & test & 3610 & \citet{ostermann_mcscript20_2019} \\
MCTest & test & 840 & \citet{richardson_mctest_2013} \\
QASC & validation & 926 & \citet{khot_qasc_2020} \\
RACE & test & 4934 & \citet{lai_race_2017} \\
RACE-C & test & 712 & \citet{liang_new_2019} \\
SciQ & test & 884 & \citet{welbl_crowdsourcing_2017} \\ \hline
\end{tabular}
\caption{Datasets used for the multiple choice setting, including split used and sample size. Validation splits were used for CosmosQA since the test split is not publicly available, and for QASC since context passages or gold answers are not available.}
\label{tab:mc_qa_datasets}
\end{table*}

\begin{table*}[]
\centering
\begin{tabular}{lll}
\hline
Dataset & Size & Reference \\ \hline
BioASQ & 1504 & \citet{fisch_mrqa_2019} \\
TriviaQA & 7785 &  \\
HotpotQA & 5901 &  \\
SQuAD & 10506 &  \\
Natural Questions & 12836 &  \\
SQuAD2 & 11871 & \citet{rajpurkar_know_2018} \\
SQuAD-adv & 5347 & \citet{jia_adversarial_2017} \\ \hline
\end{tabular}
\caption{Extractive QA datasets used. Validation sets are used on the SQuAD2.0 and SQuAD adversarial datasets. MRQA 2019 dev sets are used for the other five datasets.}
\label{tab:extractive_qa_datasets}
\end{table*}

\section{QA models}
\label{sec:appendix_qa_models}

\begin{table*}[]
\centering
\begin{tabular}{lll}
\hline
Hugging Face & Name \\ \hline
LIAMF-USP/roberta-large-finetuned-RACE & RoBERTa-RACE \\
bert-large-uncased-whole-word-masking-finetuned-squad & BERT-Large \\
distilbert-base-uncased-distilled-squad & DistillBERT \\
ynie/albert-xxlarge-v2-snli\_mnli\_fever\_anli\_R1\_R2\_R3-nli & albert-anli \\
microsoft/deberta-base-mnli & mnli-base \\
microsoft/deberta-v2-xxlarge-mnli & mnli-large \\ \hline
\end{tabular}
\caption{Pretrained QA and NLI models used.}
\label{tab:models_trained1}
\end{table*}

\begin{table*}[]
\centering
\begin{tabular}{lllll}
\hline
Model & Dataset & Epochs & Score &  \\ \hline
t5-small & \citet{demszky_transforming_2018} & 20 & Rogue1 & 90.73 \\
deberta-v3-xsmall & \citet{welbl_crowdsourcing_2017} & 6 & Accuracy & 93.99 \\
deberta-v3-base & \citet{welbl_crowdsourcing_2017} & 6 & Accuracy & 91.79 \\ \hline
\end{tabular}
\caption{The models we trained for or setups with evaluation scores and number of epochs trained.}
\label{tab:models_trained2}
\end{table*}

Table \ref{tab:models_trained1} outlines the pretrained QA models that we used and the datasets they are trained on. All these models are publicly available on the Hugging Face hub under the locations listed. Where space doesn't allow, RoBERTa-RACE is aliased as RACE.

We trained the two DeBERTa-v3 models (xsmall and base) as shown in Table \ref{tab:models_trained2}. They were trained using the Hugging Face trainer API \citep{wolf_huggingfaces_2020} with an Adam optimizer at a learning rate of 5.60e-05 with weight decay of 0.01. All models and inference were performed on 1 Tesla P100 GPU. Full instructions on reproducibility as well as trained models are provided in the publicly available code, including directions to weights and biases to inspect the training runs, full parameter set, and evaluation suites.

\section{QA2D models}
\label{sec:appendix_qa2d_models}

A QA2D model reformulates a question-answer pair to a declarative statement \citep{demszky_transforming_2018}. As noted in \citet{chen_can_2021} and \citet{mishra_looking_2021}, the QA2D reformulation is critical to using NLI models in QA since the proposed answer needs to match the format of NLI. We trained a T5-small model \citep{raffel_exploring_2020} on the dataset proposed by \citet{demszky_transforming_2018} for QA2D since we found almost no noticeable differences in performance in larger models. This used the same setup as the DeBERTa-v3 models xsmall and base (see Table \ref{tab:models_trained2}).

Unlike \citet{chen_can_2021}, we found that regardless of size, these QA2D models struggled with long questions or questions with complex syntax and would often leave the answer out of the statement. In order to solve this, constrained decoding that required the answer to be in the statement was tried. However, this often produced ungrammatical or nonsensical statements. We settled with the following heuristic to postprocess QA2D outputs: If less than 50\% of the tokens in the answer were in the statement then we appended the answer to the end of the statement. 50\% was used to account for rephrasing the answer or swapping pronouns. While some statements resulted in answer redundancy, this was better than having hypotheses which left out the answer. 

Future work on QA2D should focus on how these models can be used outside of the domains in the dataset provided by \citet{demszky_transforming_2018}. Finally it is important to note that erroneous QA2D outputs could effect the quality of the whole pipeline see \citet{chen_can_2021} for a more detailed analysis of this.

\section{NLI models}
\label{sec:appendix_nli_models}

NLI is used to classify whether the reformulated answer is contradicted, entailed, or neutral with respect to a context passage. We used the whole context, as \citet{schuster_stretching_2022} and \citet{mishra_looking_2021} demonstrated that long premises still performed adequate though not as well as sentence-length premises. Using the whole context avoids needing to use decontextualization as is required in \citet{chen_can_2021}. 

We used two DeBERTa-based models \citep{he_deberta_2021} trained on the MNLI dataset \citep{williams-etal-2018-mnli} (called mnli-base and mnli-large) and an ALBERT model \citep{albert} trained on the ANLI dataset in addition to various other NLI datasets (called albert-anli) \citep{nie_adversarial_2020}. Table \ref{tab:models_trained1} contains the Hugging Face references to the NLI models After inference, the confidence scores are used for answer selection and performance evaluation.

\subsection{Model size and approach performance analysis}
\label{sec:appendix_model_size}

\begin{table*}[]
\centering
\begin{tabular}{llllllllllll}
\hline
{\small QA Model} & {\small Cosmos} &  {\small DREAM} & {\small MCS} & {\small MCS2} & {\small MCT} & {\small QASC} & {\small RACE} & {\small RACE-C} & {\small SciQ} & {\small \emph{Average}} \\ \hline
SciQ-base & 18.46 & 43.80 & 61.99 & 63.71 & 44.76 & 93.41 & 30.97 & 27.39 & 95.28 & 53.31 \\
SciQ-small &  25.46 & 48.26 & 60.28 & 66.04 & 59.76 & 90.60 & 35.56 & 30.62 & 98.09 & 57.19 \\
RACE & 64.22 & 82.56 & 89.70 & 86.98 & 90.48 & 98.16 & 76.93 & 69.80 & 97.96 & 84.09 \\ \hline
mnli-large \\
E+C & 44.36 & 80.94 & 85.52 & 84.99 & 90.60 & 96.44 & 64.29 & 51.40 & 92.47 & 76.77 \\
E & 36.18 & 79.03 & 86.02 & 79.72 & 89.88 & 95.90 & 62.14 & 49.72 & 91.96 & 74.50 \\
C & 59.26 & 78.98 & 83.12 & 84.43 & 89.29 & 92.76 & 62.74 & 47.05 & 91.58 & 76.58 \\ \hline
mnli-base \\ 
 QA + E + C & 64.32 & 82.66 & 89.63 & 87.01 & 90.71 & 98.27 & 76.95 & 69.80 & 98.09 & 84.16 \\
  QA + E & 64.25 & 82.66 & 89.63 & 86.98 & 90.71 & 98.27 & 76.95 & 69.80 & 97.96 & 84.14 \\
  QA + C & 64.29 & 82.56 & 89.63 & 87.01 & 90.60 & 98.16 & 76.93 & 69.80 & 97.96 & 84.1 \\
  E + C & 33.03 & 62.27 & 76.76 & 72.11 & 68.57 & 92.66 & 45.16 & 34.41 & 88.01 & 63.66 \\
  E & 27.81 & 62.47 & 79.37 & 71.94 & 68.81 & 92.66 & 43.48 & 34.41 & 88.01 & 63.22 \\
  C & 43.45 & 59.19 & 70.18 & 69.97 & 67.50 & 81.86 & 41.81 & 32.58 & 87.37 & 61.55 \\ \hline
 albert-anli \\ 
QA + E + C & 64.19 & 82.56 & 89.70 & 87.06 & 90.48 & 98.16 & 76.93 & 69.80 & 97.96 & 84.09 \\
  QA + E & 64.19 & 82.56 & 89.70 & 87.06 & 90.60 & 98.16 & 76.93 & 69.80 & 97.96 & 84.11 \\
  QA + C & 64.22 & 82.56 & 89.70 & 86.98 & 90.48 & 98.16 & 76.93 & 69.80 & 97.96 & 84.09 \\
  E + C & 35.71 & 68.20 & 79.55 & 73.88 & 77.50 & 91.79 & 49.05 & 39.47 & 90.82 & 67.33 \\
  E & 33.67 & 68.35 & 79.91 & 73.19 & 77.38 & 91.90 & 49.07 & 39.19 & 90.94 & 67.07 \\
  C & 45.16 & 63.74 & 73.58 & 72.71 & 73.33 & 77.86 & 46.34 & 38.20 & 87.24 & 64.24 \\ \hline
\end{tabular}
\caption{Accuracy scores in the multiple choice setting for various NLI models used. Calibration was with the RoBERTA-RACE model.}
\label{tab:cross_mc_performance}
\end{table*}

Table \ref{tab:cross_mc_performance} mirrors Table \ref{tab:calibrated_performance} in the main text, but shows the accuracy results for uncalibrated \textbf{E}, \textbf{C}, and \textbf{E+C} in the main mnli-large model, as well as the results with the other NLI models, mnli-base and albert-anli. Table \ref{tab:selective_ranking} shows selective QA accuracy in the multiple choice setting where answer selection is done through ranking before we rank answers for selective QA. Selective QA on extractive QA using DistillBERT (table \ref{tab:selective_distillbert}) shows that \textbf{QA+E+C} does best in all cases and contradiction only does second best at 50\% coverage.

\begin{table*}[]
\centering
\begin{tabular}{lllllllll}
\hline
 & Dataset & QA+E+C & QA+E & QA+C & E+C & E & C & QA \\ \hline
20\% & CosmosQA & 77.55 & 67.17 & {\ul 83.25} & 20.10 & 27.47 & 67.50 & \textbf{88.61} \\
 & DREAM & \textbf{98.28} & 96.32 & {\ul 96.81} & 81.13 & 91.91 & 93.87 & \textbf{98.28} \\
 & MCScript & \textbf{99.82} & \textbf{99.64} & {\ul 99.46} & 93.02 & {\ul 98.93} & 96.96 & \textbf{99.82} \\
 & MCScript-2.0 & \textbf{99.58} & {\ul 99.03} & 97.37 & 92.24 & 97.37 & 95.01 & \textbf{99.58} \\
 & MCTest & \textbf{100} & \textbf{100} & {\ul 99.40} & 85.12 & 97.02 & 97.02 & 98.81 \\
 & QASC & \textbf{100} & \textbf{100} & \textbf{100} & 97.30 & \textbf{100} & {\ul 99.46} & \textbf{100} \\
 & RACE & {\ul 94.93} & 92.13 & 90.17 & 62.73 & 76.71 & 75.05 & \textbf{98.24} \\
 & RACE-C & {\ul 88.73} & 85.21 & 86.62 & 71.13 & 74.65 & 69.01 & \textbf{93.66} \\
 & SciQ & \textbf{100} & \textbf{100} & \textbf{100} & 82.05 & 100 & 96.15 & \textbf{100} \\
 & Avg & {\ul 95.43} & 93.28 & {\ul 94.79} & 76.09 & 84.90 & 87.78 & \textbf{97.45} \\ \hline
50\% & CosmosQA & {\ul 80.29} & 70.78 & \textbf{80.70} & 32.17 & 34.72 & 64.88 & 76.47 \\
 & DREAM & {\ul 95.10} & 93.63 & 93.63 & 85.20 & 89.41 & 88.33 & \textbf{96.67} \\
 & MCScript & \textbf{98.57} & {\ul 97.85} & 97.14 & 94.71 & 95.99 & 92.70 & \textbf{98.78} \\
 & MCScript-2.0 & {\ul 96.40} & 94.46 & {\ul 96.07} & 91.02 & 91.75 & 91.69 & \textbf{98.01} \\
 & MCTest & \textbf{99.52} & {\ul 98.81} & \textbf{99.76} & 91.43 & 95.24 & 96.19 & \textbf{99.52} \\
 & QASC & \textbf{100} & \textbf{99.78} & \textbf{99.78} & 98.27 & {\ul 98.70} & 98.49 & \textbf{100} \\
 & RACE & {\ul 90.11} & 87.22 & 85.23 & 67.89 & 71.70 & 68.18 & \textbf{93.88} \\
 & RACE-C & {\ul 85.11} & 78.09 & 77.25 & 66.57 & 66.85 & 55.06 & \textbf{87.36} \\
 & SciQ & \textbf{100} & \textbf{100} & \textbf{99.74} & 89.03 & 96.43 & 96.43 & \textbf{100} \\
 & Avg & {\ul 93.90} & 91.18 & 92.14 & 79.59 & 82.31 & 83.55 & \textbf{94.52} \\ \hline
\end{tabular}
\caption{Selective QA accuracies in the multiple choice setting where answer selection is done through ranking before we rank answers for selective QA.}
\label{tab:selective_ranking}
\end{table*}

\begin{table*}[]
\centering
\begin{tabular}{lllllllll}
\hline
 & Dataset & QA+E+C & QA+E & QA+C & E+C & E & C & QA \\ \hline
20\% & BioASQ & 70.97 & 70.41 & 71.55 & {\ul 74.07} & {\ul 74.07} & \textbf{74.34} & 68.99 \\
 & HotpotQA & \textbf{73.44} & {\ul 73.08} & 70.88 & 71.59 & 71.51 & 70.41 & 69.41 \\
 & Natural Questions & \textbf{85.59} & 85.29 & {\ul 85.45} & 78.46 & 78.46 & 80.53 & 83.27 \\
 & SQuAD & 96.22 & {\ul 96.45} & 95.77 & 83.15 & 83.09 & 81.37 & \textbf{97.15} \\
 & SQuAD-adv & {\ul 40.39} & 39.75 & 39.49 & 40.07 & 39.56 & \textbf{40.59} & 31.98 \\
 & SQuAD2 & 35.46 & 35.24 & 33.64 & {\ul 36.36} & 36.13 & \textbf{36.66} & 25.95 \\
 & TriviaQA & \textbf{64.96} & {\ul 64.68} & 64.55 & 52.67 & 52.09 & 52.56 & 63.98 \\
 & Avg & \textbf{66.72} & {\ul 66.41} & 65.90 & 62.34 & 62.13 & 62.35 & 62.96 \\ \hline
50\% & BioASQ & {\ul 65.96} & 65.92 & 64.37 & 63.53 & 63.53 & \textbf{66.95} & 64.79 \\
 & HotpotQA & 64.42 & 64.21 & 63.65 & 65.88 & {\ul 65.85} & \textbf{66.91} & 62.81 \\
 & Natural Questions & {\ul 72.28} & 71.99 & 70.82 & 67.54 & 67.51 & \textbf{74.18} & 69.95 \\
 & SQuAD & {\ul 92.56} & \textbf{92.57} & 92.34 & 81.86 & 82.21 & 80.95 & 92.54 \\
 & SQuAD-adv & 33.69 & 32.90 & 33.45 & \textbf{38.74} & 38.22 & {\ul 38.52} & 31.89 \\
 & SQuAD2 & 26.68 & 25.70 & 26.00 & \textbf{32.95} & 32.61 & {\ul 32.83} & 23.52 \\
 & TriviaQA & {\ul 58.40} & \textbf{58.41} & 58.25 & 51.43 & 51.18 & 52.99 & 58.25 \\
 & Avg & \textbf{59.14} & 58.81 & 58.41 & 57.42 & 57.30 & {\ul 59.05} & 57.68 \\ \hline
\end{tabular}
\caption{SelectiveQA on extractive QA using DistillBERT. Note that QA+E+C does best in all cases and contradiction only does second best at 50\% coverage.}
\label{tab:selective_distillbert}
\end{table*}

\section{Calibration models}
\label{sec:appendix_calibration_models}

Like \citet{kamath_selective_2020} and \citet{chen_can_2021} we developed a set of calibration models in order to perform answer ranking. A calibration model is trained on a set of posterior probabilities from downstream models to predict whether an answer is correct. 

To compare the effect of using different combinations of NLI class confidence scores, we trained a logistic regression model on linear combinations of the following features: \textbf{QA} indicates that the QA model confidence score is being used, \textbf{E} indicates the entailment score, \textbf{C} indicates the contradiction score, and \textbf{N} indicates the neutral score. Like in \citet{chen_can_2021}, all calibration models are trained on a holdout set of 100 samples from a single domain using logistic regression which predicts, given the confidence scores of the downstream models, whether the answer is correct. A multi-domain calibration approach like in \citet{kamath_selective_2020} was not used since the focus was a minimum experiment to test the viability of leveraging different NLI classifications. 

\subsection{Regression Analysis}
\begin{table*}[]
\centering
\begin{tabular}{lllllll}
\hline
QA Model & NLI Model & Combination & Confidence & Entailment & Contradiction & Acc \\ \hline
SciQ & mnli-base & QA + C & 4.13 &  & -1.06 & 0.99 \\
 &  & QA + E & 3.90 & 1.37 &  & 0.99 \\
 &  & QA + E + C & 3.83 & 1.22 & -0.76 & 0.99 \\
 &  & E + C &  & 2.56 & -1.47 & 0.86 \\
 & mnli-large & QA + C & 3.98 &  & -1.32 & 0.99 \\
 &  & QA + E & 3.78 & 1.55 &  & 0.99 \\
 &  & QA + E + C & 3.65 & 1.31 & -0.97 & 0.99 \\
 &  & E + C &  & 2.63 & -1.72 & 0.91 \\
RACE & mnli-base & QA + C & 3.04 &  & -0.15 & 0.89 \\
 &  & QA + E & 3.03 & 0.27 &  & 0.89 \\
 &  & QA + E + C & 3.02 & 0.26 & -0.14 & 0.89 \\
 &  & E + C &  & 0.73 & -0.46 & 0.75 \\
 & mnli-large & QA + C & 2.97 & 0.00 & -0.81 & 0.89 \\
 &  & QA + E & 2.91 & 0.98 &  & 0.89 \\
 &  & QA + E + C & 2.85 & 0.92 & -0.75 & 0.89 \\
 &  & E + C &  & 1.76 & -1.12 & 0.78 \\ \hline
\end{tabular}
\caption{Regression analysis for each mnli-based nli model with each QA model used calibration with logistic regression multiple choice settings. Accuracy is the evaluation metric used.}
\label{tab:regression_analysis}
\end{table*}

To illustrate the characteristics of the calibration models, we present a regression analysis for the multiple choice setting (Table \ref{tab:regression_analysis}). The results indicate that as the mnli model gets larger, the calibration model uses its NLI confidence scores more. Importantly, entailment coefficients are stronger than contradiction coefficients in all cases.

\section{Correlation Analysis}
\label{sec:appendix_correlation_analysis}

\begin{table*}[]
\centering
\begin{tabular}{llllllll}
\hline
 &  & Contradiction &  & Entailment &  & Neutral &  \\ \hline
Dataset & QA & Score & Class & Score & Class & Score & Class \\
CosmosQA & \textbf{0.53} & {\ul -0.34} & -0.17 & 0.05 & -0.01 & 0.21 & 0.16 \\
DREAM & \textbf{0.72} & {\ul -0.57} & -0.35 & 0.54 & 0.50 & -0.11 & -0.13 \\
MCScript & \textbf{0.80} & {\ul -0.59} & -0.42 & {\ul 0.59} & 0.50 & -0.04 & -0.08 \\
MCScript2 & \textbf{0.77} & {\ul -0.50} & -0.32 & 0.41 & 0.37 & -0.04 & -0.05 \\
MCTest & \textbf{0.73} & -0.65 & -0.47 & 0.64 & {\ul 0.69} & -0.20 & -0.15 \\
QASC & {\ul 0.57} & -0.54 & -0.28 & 0.55 & \textbf{0.67} & -0.50 & -0.26 \\
RACE & \textbf{0.65} & {\ul -0.37} & -0.20 & 0.35 & 0.34 & -0.11 & -0.11 \\
RACE-C & \textbf{0.59} & -0.24 & -0.13 & 0.18 & {\ul 0.25} & -0.09 & -0.11 \\
SciQ & \textbf{0.75} & {\ul -0.69} & -0.47 & 0.68 & 0.67 & -0.42 & -0.19 \\ \hline
\end{tabular}
\caption{Correlation analysis (Spearman rank correlation) per dataset in the multiple choice setting. RoBERTa-RACE is used for the QA scores.}
\label{tab:per_dataset_correlation}
\end{table*}

\begin{table*}[]
\centering
\begin{tabular}{llllll}
\hline
 &  & Contradiction & Entailment & Neutral & QA \\ \hline
multiple choice & Score & \textbf{-0.47} & 0.37 & -0.06 & 0.71 \\
 & Class & -0.28 & \textbf{0.38} & -0.06 &  \\
extractive QA & Score & -0.16 & \textbf{0.31} & -0.12 & 0.19 \\
 & Class & -0.15 & \textbf{0.39} & -0.29 &  \\ \hline
\end{tabular}
\caption{Correlation analysis (Spearman rank correlation) in the multiple choice and extractive QA settings. RoBERTa-RACE is the QA model used for multiple choice QA scores and BERT-large is used for the extractive QA scores.}
\label{tab:correlation}
\end{table*}

Since we are using the NLI and QA model scores to construct the setups above, it is useful to know how these factors correlate with the correct answer. Table \ref{tab:correlation} shows how each NLI class correlates both by score and by actual classification (score > 50\%) as compared against QA model confidence score. The multiple choice analysis shows answers from the RoBERTa-RACE model and the extractive QA analysis shows answers from the BERT-large model trained on SQuAD. The correlation analysis presents Spearman rank correlations. 

What we see is that in the multiple choice setting, the confidence score has a strong correlation with the correct answer, which makes sense given the confidence score is a softmax over the multiple choice classes. Extractive QA confidence scores have a much weaker correlation and tend to have less correlation than entailment has with the correct answer. Despite the results presented above, contradiction only has a notable correlation with the correct answer when the score is used rather than the classification. This is a point in favor of our approach of using confidence scores for NLI rather than classifications. 

Interestingly, in the extractive QA case, the neutral class is more negatively correlated when selecting for contradiction when using classification. Our conjecture would be that in the extractive QA case, we don’t have much to compare against. When looking at the per dataset correlations for the multiple choice setting (Table \ref{tab:per_dataset_correlation}), we see that in most cases, other than the QA confidence scores, the contradiction scores have the strongest correlations with the correct answer out of any NLI class and neutral, as we would expect, tends to have very weak correlations. We do not present the per dataset correlation for extractive QA as they are very weak, which we again hypothesize comes from having no answers to compare with.

\end{document}